\documentclass[11pt,a4paper]{article}
\usepackage[hyperref]{acl2021}
\usepackage{times}
\usepackage{latexsym}

\usepackage{amsmath,amsfonts,bm}

\def\eqref#1{equation~\ref{#1}}

\def\1{\bm{1}}

\def\vv{{\bm{v}}}

\def\vx{{\bm{x}}}
\def\vy{{\bm{y}}}

\DeclareMathAlphabet{\mathsfit}{\encodingdefault}{\sfdefault}{m}{sl}
\SetMathAlphabet{\mathsfit}{bold}{\encodingdefault}{\sfdefault}{bx}{n}

\def\sR{{\mathbb{R}}}

\usepackage{microtype}
\usepackage{bbm}
\usepackage{amsmath}
\usepackage{amsfonts}
\usepackage{algorithm}
\usepackage{algpseudocode}
\usepackage{booktabs}
\usepackage{xparse}
\usepackage{graphicx}
\usepackage{multicol}
\usepackage{multirow}
\usepackage{tikz}
\usetikzlibrary{shapes,arrows}
\usepackage{natbib}
\usepackage{caption}
\usepackage{subcaption}
\usepackage{tikz-dependency}
\usepackage{wrapfig}
\usepackage{pgfplots}
\pgfplotsset{compat=1.17} 
\usepackage[T1]{fontenc}
\usepackage{subfiles}
\usepackage{readarray}
\usepackage{xcolor}
\usepackage{import}
\usepackage{hhline}
\usepackage{enumitem}
\usepackage{boldline}
\usepackage{mathtools}
\usepackage{colortbl}

\aclfinalcopy 
\def\aclpaperid{1229} 

\newcommand{\xvec}{\mathbf{x}}

\newcommand{\rvec}{\mathbf{r}}

\newcommand{\mcL}{\mathcal{L}}

\newcommand{\mcY}{\mathcal{Y}}

\newcommand{\Wvec}{\mathbf{W}}

\newcommand{\bvec}{\mathbf{b}}

\title{Improving Named Entity Recognition by \\ External Context Retrieving and Cooperative Learning}

\author{Xinyu Wang$^{\diamond\ddagger}$, Yong Jiang$^{\dagger}$\textsuperscript{$\ast$}, Nguyen Bach$^{\dagger}$, Tao Wang$^{\dagger}$,\\
\textbf{Zhongqiang Huang$^{\dagger}$, Fei Huang$^{\dagger}$,  Kewei Tu$^{\diamond}$}\thanks{\hspace{1mm} Yong Jiang and Kewei Tu are the corresponding authors. $^{\ddagger}$: This work was conducted when Xinyu Wang was interning at Alibaba DAMO Academy. } \\
 $^\diamond$School of Information Science and Technology, ShanghaiTech University \\
 Shanghai Engineering Research Center of Intelligent Vision and Imaging \\
 Shanghai Institute of Microsystem and Information Technology, Chinese Academy of Sciences \\
 University of Chinese Academy of Sciences \\
 $^\dagger$DAMO Academy, Alibaba Group \\
  {\tt \{wangxy1,tukw\}@shanghaitech.edu.cn, yongjiang.jy@alibaba-inc.com} \\
  {\tt \{nguyen.bach,leeo.wangt,z.huang,f.huang\}@alibaba-inc.com} \\
}
\date{}

\begin{document}
\maketitle
\begin{abstract}

Recent advances in Named Entity Recognition (NER) show that document-level contexts can significantly improve model performance. In many application scenarios, however, such contexts are not available. In this paper, we propose to find external contexts of a sentence by retrieving and selecting a set of semantically relevant texts through a search engine, with the original sentence as the query. We find empirically that the contextual representations computed on the retrieval-based input view, constructed through the concatenation of a sentence and its external contexts, can achieve significantly improved performance compared to the original input view based only on the sentence. Furthermore, we can improve the model performance of both input views by Cooperative Learning, a training method that encourages the two input views to produce similar contextual representations or output label distributions. Experiments show that our approach can achieve new state-of-the-art performance on 8 NER data sets across 5 domains.\footnote{Our newest code is publicly available at \url{https://github.com/modelscope/AdaSeq/tree/master/examples/RaNER}. The older version: \url{https://github.com/Alibaba-NLP/CLNER}.}

\end{abstract}

\section{Introduction}
Pretrained contextual embeddings such as ELMo \citep{peters-etal-2018-deep}, Flair \citep{akbik-etal-2018-contextual} and BERT \citep{devlin-etal-2019-bert} have significantly improved the accuracy of Named Entity Recognition (NER) models. Recent work \citep{devlin-etal-2019-bert,yu-etal-2020-named,yamada-etal-2020-luke} found that including document-level contexts of the target sentence in the input of contextual embeddings methods can further boost the accuracy of NER models. 
However, there are a lot of application scenarios in which document-level contexts are unavailable in practice. For example, there are sometimes no available contexts in users' search queries, tweets and short comments in various domains such as social media and E-commerce domains. When professional annotators annotate ambiguous named entities in such cases, they usually rely on domain knowledge for disambiguation. This kind of knowledge can often be found through a search engine. Moreover, when the annotators are not sure about a certain entity, they are usually encouraged to find related knowledge through a search engine \citep{wang-etal-2019-crossweigh}. Therefore, we believe that NER models can benefit from such a process as well.

\begin{figure}[t]
	\centering
	\includegraphics[scale=0.215]{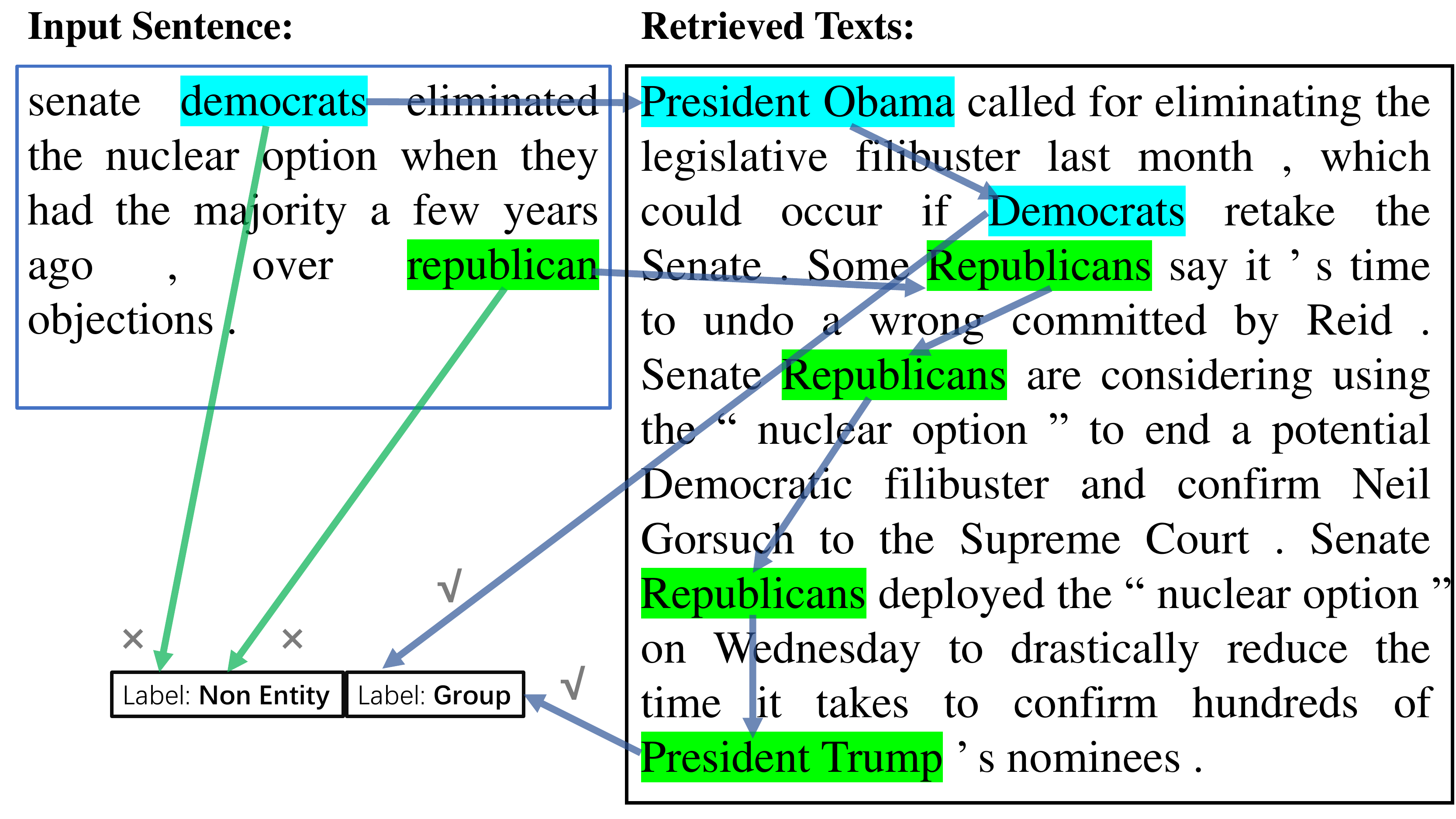}
	\caption{A motivating example from WNUT-17 dataset. The retrieved texts help the model to correctly predict the named entities of ``democrats'' and ``republican''.}
	\label{fig:motivate}
\end{figure}

In this paper, we propose to improve NER models by retrieving texts related to the input sentence by an off-the-shelf search engine. We re-rank the retrieved texts according to their semantic relevance to the input sentence and select several top-ranking texts as the external contexts. Consequently, we concatenate the input sentence and external contexts together as a new retrieval-based input view and feed it to the pretrained contextual embedding module, so that the resulting semantic representations of the input tokens can be improved. The token representations are then fed into a CRF layer for named entity prediction. A motivating example is shown in Figure \ref{fig:motivate}.

Moreover, we consider utilizing the new input view to improve model performance with the original input view that does not have external contexts. This can be useful in application scenarios when external contexts are unavailable or undesirable (e.g., in time-critical scenarios). To this end, we propose Cooperative Learning (CL) that encourages the two input views to produce similar predictions. We propose two approaches to CL which minimize either the $L_2$ distances between the token representations of the two input views or the Kullback–Leibler (KL) divergence between the prediction distributions of the two input views during training. 

Our experiments show that including the retrieved external contexts can significantly improve the accuracy of NER models on 8 NER datasets from 5 domains. With CL, the accuracy of the NER models with both input views can be further improved. Our approaches outperform previous state-of-the-art approaches in each domain. 

The contributions of this paper are:
\begin{enumerate}[leftmargin=*]
    \item We propose a simple and straight-forward way to improve the contextual representation of an input sentence through retrieving related texts using a search engine. We take the retrieved texts together with the input sentence as a new retrieval-based view.
    \item 
    We propose Cooperative Learning to jointly improve the accuracy of both input views in a unified model. We propose two approaches in CL based on the $L_2$ norm and KL divergence respectively. CL can utilize unlabeled data for further improvement.
    \item We show the effectiveness of our approaches in several NER datasets across 5 domains and our approaches achieve state-of-the-art accuracy. By leveraging a large amount of unlabeled data, the performance can be further improved. 
\end{enumerate}
\begin{figure*}[ht]
	\centering
	\includegraphics[scale=0.45]{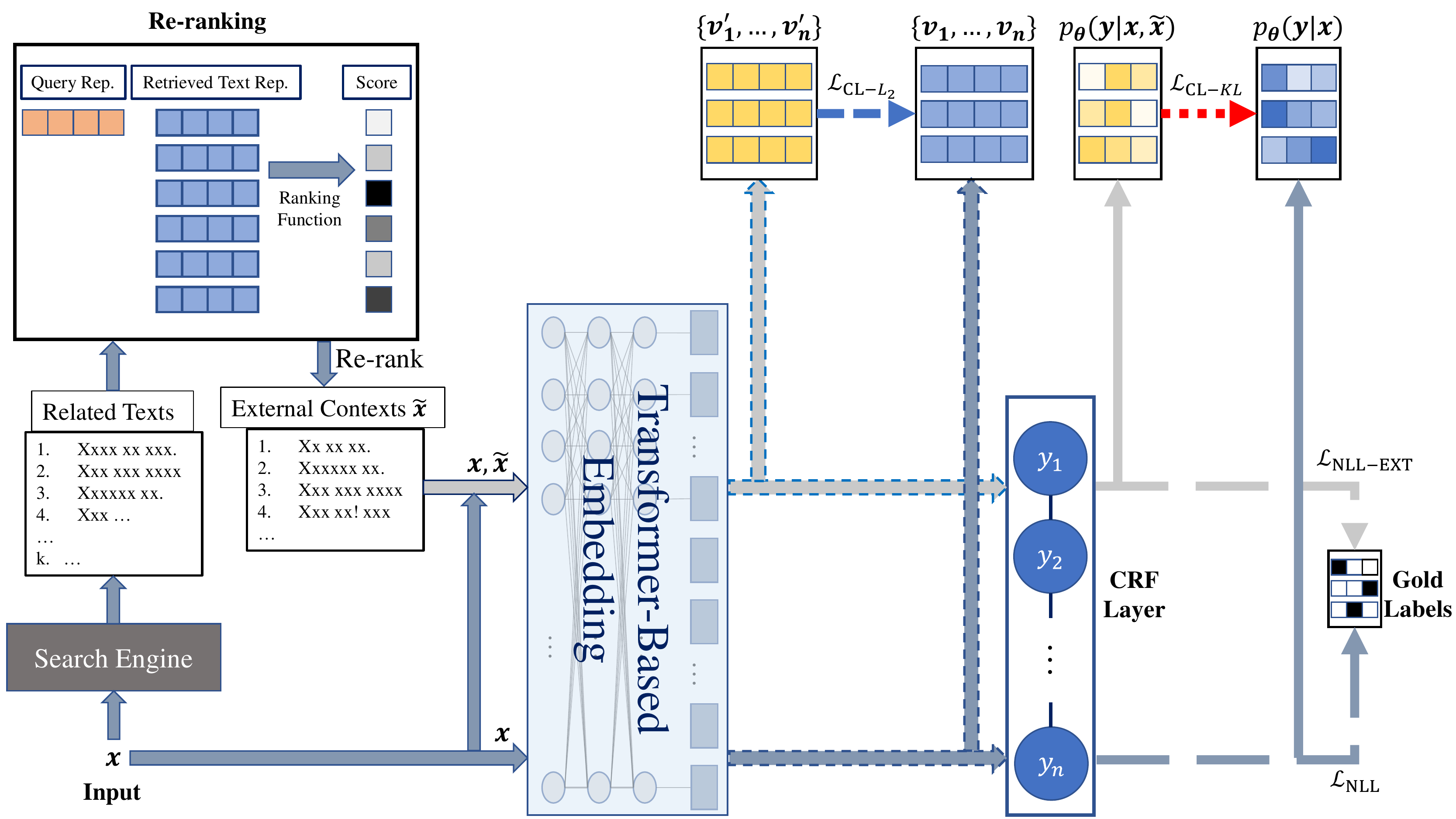}
	\caption{The architecture of our framework. An input sentence $\xvec$ is fed into a search engine to get $k$ related texts. The related texts are then fed into the re-ranking module. The framework selects $l$ highest ranking related texts output from the re-ranking module and feeds the texts to a transformer-based model together with the input sentence. Finally, we calculate the negative likelihood loss $\mcL_{\text{NLL}}$ and $\mcL_{\text{NLL-EXT}}$ together with the CL loss (either $\mcL_{\text{CL-}L_2}$ or $\mcL_{\text{CL-KL}}$).}
	\label{fig:architecture}
\end{figure*}

\section{Framework}
Given a sentence of $n$ tokens $\vx = \{x_1, \cdots, x_n\}$, the input sentence is fed into a search engine as a query. The search engine returns the top $k$ relevant texts $\{\hat{\vx}_1, \cdots, \hat{\vx}_k\}$. Our framework feeds these texts into a re-ranking model. We concatenate $l$ top-ranking texts output from the re-ranking model as the external contexts. The NER model is fed with either an input view with the input sentence (original input view) or a concatenation of the input sentence and external contexts (retrieval-based input view) as input. The model outputs the predictions of labels $\vy = \{y_1, \cdots, y_n\}$ at each position based on the CRF layer. To further improve the model, we use Cooperative Learning to train a unified model that is strong in both input views. With CL, the model is additionally constrained to be consistent in the internal representations or the output distributions of both input views. The architecture of our framework is shown in Figure \ref{fig:architecture}.

\subsection{Re-ranking}
Given an input sentence as a search query, the search engine returns ranked relevant texts. 
However, the off-the-shelf search engine is highly optimized for a fast speed over a large set of documents, so it may sometimes produce semantically irrelevant results or rank the results using inaccurate relevance scores.
Since the NER task targets at semantically recognizing named entities, it is more helpful if the relevant texts are semantically similar to the input sentence. Therefore, we need to re-rank the retrieved texts so that the most semantically relevant texts are chosen. We propose to apply BERTScore \citep{Zhang*2020BERTScore:} to score the relatedness of each retrieved text to the input sentence. BERTScore is a language generation metric that calculates a sum of cosine similarity between token representations of two sentences. Therefore, it is more likely that the search query and the retrieved texts have strong semantic relations when BERTScore is large. The token representations are generated from pretrained contextual embeddings such as BERT. Given the corresponding pre-normalized token representations $\{\rvec_1, \cdots, \rvec_n\}$ of the input sentence $\vx$ and the pre-normalized token representations $\{\hat{\rvec}_1, \cdots, \hat{\rvec}_m\}$ of a certain retrieved text $\hat{\vx}$ with $m$ words, the Precision (P), Recall (R) of BERTScore measure the semantic similarities from one to another:
\begin{align*} 
    &\text{R} =\frac{1}{n} \sum_{x_i \in \vx}   \max_{\hat{x}_j \in \hat{\vx}} \rvec_i^\top \hat{\rvec}_{j};\;\;
    \text{P} = \frac{1}{m}  \sum_{\hat{x}_j\in \hat{\vx}}   \max_{x_i\in \vx}  \rvec_{i}^\top \hat{\rvec}_{j}
\end{align*}
 We re-rank the retrieved texts by the F1 scores $\text{F1}{=} 2\frac{\text{P} \cdot \text{R} }{\text{P} + \text{R} }$ and concatenate $l$ top-ranking texts $\{\hat{\vx}_1, \cdots, \hat{\vx}_l\}$ with F1 scores together as the external contexts:
\begin{displaymath}
\tilde{\vx} = [sep\_token; \hat{\vx}_1; \cdots; \hat{\vx}_l]
\end{displaymath} 
where $sep\_token$ is a special token representing a separate of sentences in the transformer-based pretrained contextual embeddings (for example, ``[SEP]'' in BERT).

\subsection{NER Model}
We solve the NER task as a sequence labeling problem. We apply a neural model with a CRF layer, which is one of the most popular state-of-the-art approaches to the task \citep{lample-etal-2016-neural,ma-hovy-2016-end,akbik-etal-2019-pooled}. In the sequence labeling model, the input sentence $\vx$ is fed into a transformer-based pretrained contextual embeddings model to get the token representations $\{\vv_1, \cdots, \vv_n\}$ by $\vv_i {=} \text{embed}_i (\vx)$.
The token representations are fed into a CRF layer to get the conditional probability $p_\theta(\vy|\vx)$:
\begin{align}
    \psi(y', y, \vv_i) &= \exp(\Wvec_{y}^{T} \vv_i + \bvec_{y',y}) \label{eq:psi}\\
    p_\theta(\vy|\vx) &= \frac{\prod\limits_{i=1}^{n} \psi(y_{i-1}, y_i, \vv_i)}{\sum\limits_{\vy' \in \mathcal{Y}(\vx)} \prod\limits_{i=1}^{n} \psi(y'_{i-1}, y'_i, \vv_i)}\nonumber
\end{align}
where $\psi$ is the potential function and $\theta$ represents the model parameters. $\mathcal{Y}(\vx)$ denotes the set of all possible label sequences given $\vx$. $y_0$ is defined to be a special start symbol. $\Wvec^{T}\in \sR^{t\times d}$ and $\bvec \in \sR^{t \times t}$ are parameters computing emission and transition scores respectively. $d$ is the hidden size of $\vv$ and $t$ is the size of the label set. During training, the negative log-likelihood loss for the input sequence with gold labels $\vy^*$ is defined by:
\begin{align}
\mcL_{\text{NLL}}(\theta) = - \log p_\theta(\vy^*|\vx) \label{eq:nll_loss}
\end{align}

\begin{figure}[t]
	\centering
	\includegraphics[scale=0.28]{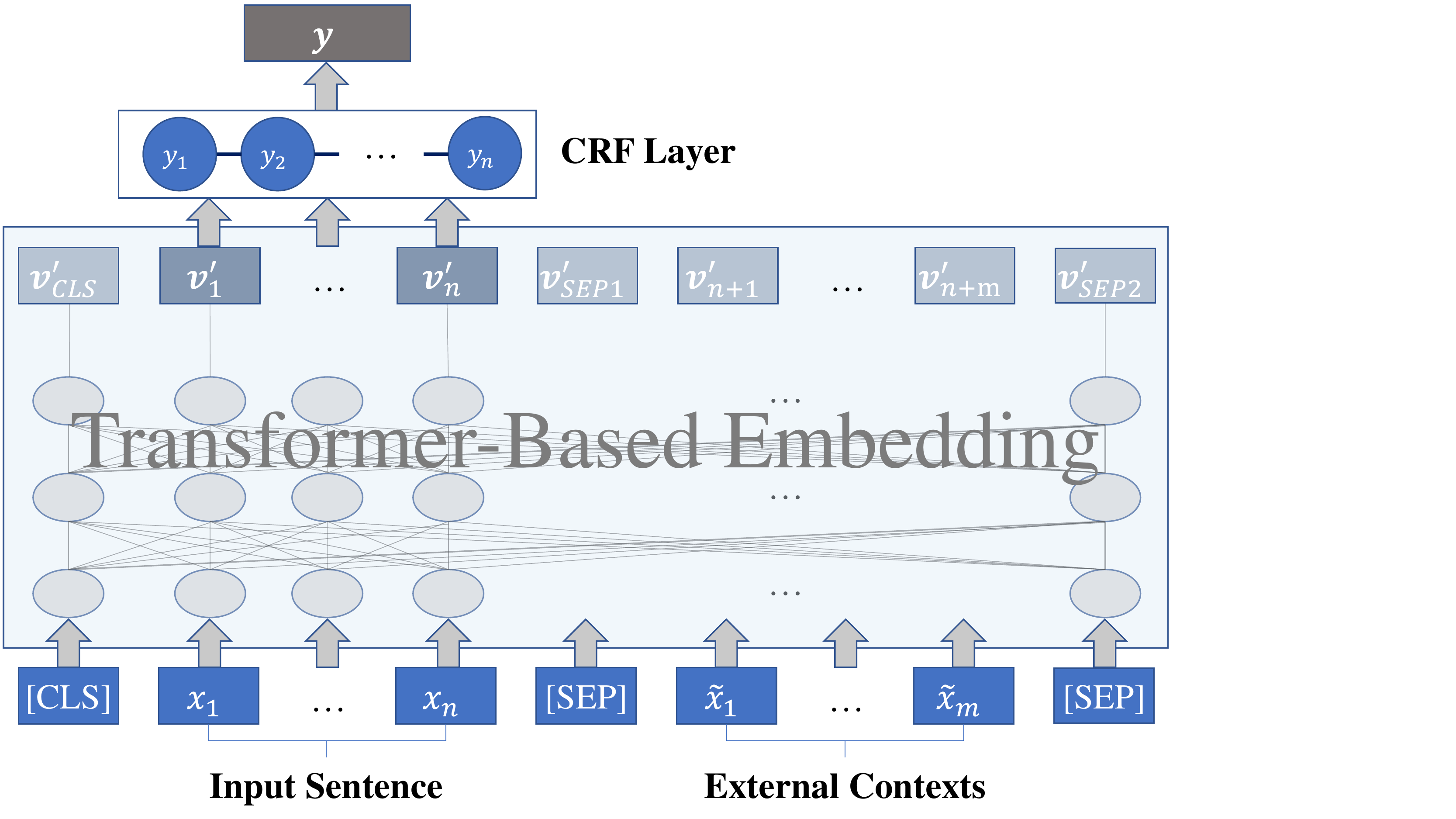}
	\caption{An illustration of our NER model architecture. ``[CLS]'' and ``[SEP]'' are an example of cls token and sep token in the embedding. }
	\label{fig:ner}
\end{figure}

In our approach, we concatenate the external contexts $\tilde{\vx}$ at the end of the input sentence $\vx$ to form the retrieval-based input view. 
The token representations are now given by:
\begin{displaymath}
\{\vv_1^{\prime},\cdots,\vv^{\prime}_n,\cdots\} = \text{embed} ([\vx;\tilde{\vx}])
\end{displaymath}
The architecture of our NER model is shown in Figure \ref{fig:ner}. Now the conditional probability $p_\theta(\vy|\vx)$ becomes $p_\theta(\vy|\vx, \tilde{\vx})$. 
The loss function in Eq. \ref{eq:nll_loss} becomes:
\begin{align}
\mcL_{\text{NLL-EXT}}(\theta) = - \log p_\theta(\vy^*|\vx,\tilde{\vx}) \label{eq:nll_loss_x}
\end{align}


\subsection{Cooperative Learning}
In practice, there are two application scenarios for the NER model: 1) offline prediction, which requires high accuracy of the prediction but the prediction speed is less emphasized; 2) online serving, which requires a faster prediction speed. The retrieval-based input view meets the requirement of the first scenario for its strong token representations. However, it does not meet the requirement of the second scenario. The external contexts are usually significantly longer than the input sentence and a search engine may not meet the latency requirements. These two issues significantly slow down the prediction speed of the model. Therefore, it is essential to improve the accuracy of the original input views in a unified model to meet these two scenarios.

Cooperative Learning targets at using the retrieval-based input view to help improve the accuracy of the model when there are no external contexts available. CL adds constraints between the internal representations or the output distributions between two input views to enforce that the predictions of both views should be near. The objective function of CL is calculated by:
\begin{align}
\mcL_{\text{CL}}(\theta)&=D(h([\vx;\tilde{\vx}]), h([\vx])) \label{eq:cl_loss}
\end{align}
where $D$ is a distance function between a function $h$ with different inputs. Because the representations or the distributions with retrieval-based input view are usually informative, we do not backpropagate the gradient through $h([\vx;\tilde{\vx}])$. We propose two approaches for CL.

\paragraph{Token Representations:} Stronger token representations usually lead to better accuracy on the task. Therefore, CL constrains the token representations of two input views to be similar. This helps the model learn to predict the token representations with external contexts even if the contexts are not available. In this approach, $D$ is the $L_2$ norm to represent the distances of the token representations:
\begin{align}
\mcL_{\text{CL-$L_2$}}(\theta)&=\sum_{i=1}^n||\vv_{i}^{\prime}-\vv_{i}||_2^2 \label{eq:L2_loss}
\end{align}

\paragraph{Label Distributions:} Since CL enforces the label predictions of both input views to be similar, a straight-forward approach is constraining the label distributions predicted by the model to be similar with the two input views. In this approach, we use the KL divergence as the function $D$. Then objective function in Eq. \ref{eq:cl_loss} becomes the KL divergence between $p_\theta(\vy|\vx, \tilde{\vx})$ and $p_\theta(\vy|\vx)$:
\begin{align}
\mcL_{\text{CL-KL}}(\theta)&{=}\sum_{\mathclap{\vy \in \mcY(\vx)}} \text{KL}(p_\theta(\vy|\vx, \tilde{\vx})||p_\theta(\vy|\vx)) \label{eq:cl_kl_loss}
\end{align}
With the CRF layer, the loss function is difficult to calculate because the output space of $p_\theta(\vy|\bullet)$ is exponential in size. To alleviate this issue, we calculate the KL divergence between the marginal distributions $q_\theta(y_i|\vx,\tilde{\vx})$ and $q_\theta(y_i|\vx)$ at each position of the sentence to approximate Eq. \ref{eq:cl_kl_loss}. The marginal distributions can be obtained using the forward-backward algorithm:
\begin{align}
\alpha(y_k)&=\sum\limits_{\{y_0,\dots,y_{k-1}\}} \prod\limits_{i=1}^{k} \psi(y_{i-1}, y_i, \vv_i)\nonumber\\
\beta(y_k) &= \sum\limits_{\{y_{k+1},\dots,y_n\}} \prod\limits_{i=k+1}^{n} \psi(y_{i-1}, y_i, \vv_i)\nonumber\\
q_\theta(y_k|\vx)
&\propto \alpha(y_k) \times \beta(y_k) \label{eq:posterior}  
\end{align}
As mentioned earlier, we do not back-propagate the gradient through $p_\theta(\vy|\vx, \tilde{\vx})$. Therefore calculating the KL divergence is equivalent to calculating the cross-entropy loss between $q(\vy|\vx, \tilde{\vx})$ and $q(\vy|\vx)$:
\begin{align}
\mcL_{\text{CL-KL}}{(}\theta{)}{=}{-}{\sum_{\mathclap{i=1}}^{n}}{\sum_{\mathclap{y_i=1}}^{t}} q_\theta(y_i|\vx,\tilde{\vx}){\log} q_\theta(y_i|\vx) \label{eq:cl_posterior_loss}
\end{align}
Together with the negative log-likelihood losses in Eq. \ref{eq:nll_loss}, \ref{eq:nll_loss_x}, the total loss in training is a summation of label losses and a CL loss:
\begin{align}
\mcL(\theta) = \mcL_{\text{NLL}}(\theta) + \mcL_{\text{NLL-EXT}}(\theta) + \mcL_{\text{CL}}(\theta) \label{eq:final_loss}
\end{align}
where $\mcL_{\text{CL}}(\theta)$ can be one of the CL loss in Eq. \ref{eq:L2_loss}, \ref{eq:cl_posterior_loss} or a summation of both of them.

\section{Experiments}

\begin{table*}[t]
\centering
\small
\begin{tabular}{l|ccccccc}
\hlineB{4}
& \textbf{\# Train} & \textbf{\# Dev} & \textbf{\# Test} & \textbf{\# Entity Labels} & \textbf{Avg. Length} & \textbf{Avg. Length w/ Context}\\
\hline
\textbf{\textsc{WNUT-16}} & 2,394 & 1,000 & 3,849 & 10 & 19.41 & 138.58 \\ 
\textbf{\textsc{WNUT-17}} & 3,394 & 1,009 & 1,287 & 6 & 18.48 & 139.49 \\
\textbf{\textsc{CoNLL-03}} & 14,987 & 3,466 & 3,684 & 4 & 13.64 & 116.23\\ 
\textbf{\textsc{CoNLL++}} & 14,987 & 3,466 & 3,466 & 4 & 13.64 & 116.23\\ 
\textbf{\textsc{BC5CDR}} & 4,560 & 4,581 & 4,797 & 2 & 25.91 & 144.13\\ 
\textbf{\textsc{NCBI}} & 5,424 & 923 & 940 & 1 & 25.01 & 135.76 \\ 
\textbf{\textsc{E-Commerce}} & 38,959 & 5,000 & 5,000 & 26 & 2.54 & 124.61\\ 
\hlineB{4}
\end{tabular}
\caption{Statistics of the dateset split, number of entity types and the average lengths with and without external contexts.}
\label{tab:stat}
\end{table*}

\subsection{Settings}
\paragraph{Datasets} To show the effectiveness of our approach, we experiment on 8 NER datasets across 5 domains:
\begin{itemize}[leftmargin=*]
    \item {\bf Social Media}: We use WNUT-16 \citep{strauss-etal-2016-results} and WNUT-17 \citep{derczynski-etal-2017-results} datasets collected from social media. We use the standard split for these datasets.
    \item {\bf News}: We use CoNLL-03 English \cite{tjong-kim-sang-de-meulder-2003-introduction} dataset and CoNLL++ \citep{wang-etal-2019-crossweigh} dataset. The CoNLL-03 dataset is the most popular dataset for NER. CoNLL++ is a revision of the CoNLL-03 datasets. \citet{wang-etal-2019-crossweigh} fixed annotation errors on the test set by professional annotators and improved the quality of the training data through their CrossWeigh approach. We use the standard dataset split for these datasets.
    \item {\bf Biomedical}: We use BC5CDR \citep{li2016biocreative} and NCBI-disease \citep{dougan2014ncbi} datasets, which are two popular biomedical NER datasets. We merge the training and development data as training set following \citet{nooralahzadeh-etal-2019-reinforcement}.
    \item {\bf Science and Technology}: We use CBS SciTech News dataset collected by \citet{jia-etal-2019-cross}. The dataset only contains the test set with the same label set as the CoNLL-03 dataset. We use the dataset to evaluate the effectiveness of cross-domain transferability from the news domain.
    \item {\bf E-commerce}: 
    We collect and annotate an internal dataset from one anonymous E-commerce website. The dataset contains 25 named entity labels for goods in short texts. We also collect 300,000 unlabeled sentences for semi-supervised training.
\end{itemize}
We show the statistics of the datasets in Table \ref{tab:stat}.

\paragraph{Annotations of the E-commerce dataset}
We manually labeled the user queries through crowdsourcing from \url{www.aliexpress.com}, which is a real-world E-commerce website. For each query, we asked one annotator to label the entities and ask another annotator to check the quality. After that, we randomly select 10\% of the dataset and ask the third annotator to check the accuracy. As a result, the overall averaged query-level accuracy\footnote{the accuracy of a query counts 1.0 if all the entities in the query are correctly recognized and 0.0 otherwise.} is 95\%. The dataset will not be released due to user privacy.

\paragraph{Retrieving and Ranking} We use an internal E-commerce search engine for the E-commerce dataset. For the other datasets, we use Google Search as the search engine. Google Search is an off-the-shelf search engine and can simulate the offline search over various domains.
We use summarized descriptions from the search results as the retrieved texts\footnote{If the descriptions are not available, we use the titles of the results instead.}. As Google Search limits the maximal length of searching queries to 32 words, we chunk a sentence into multiple sub-sentences based on punctuation if the sentence is longer than 30, feed each sub-sentence to the search engine, and retrieve up to 20 results. We filter the retrieved texts that contain any part of the datasets. Our re-ranking module selects top $6$ relevant texts\footnote{We determined that $6$ is a reasonable number based on preliminary experiments.} as the external contexts of the input sentence and chunk the external contexts if the total sub-token lengths of the input sentence and external contexts exceeds 510.

\paragraph{Model Configurations}
For the re-ranking module, we use Roberta-Large \citep{liu2019roberta} for token representations which is the default configuration in the code\footnote{\url{https://github.com/Tiiiger/bert_score}} of BERTScore \citep{Zhang*2020BERTScore:}. For token representations in the NER model, we use pretrained Bio-BERT \citep{lee2020biobert} for datasets from the biomedical domain and use XLM-RoBERTa \citep{conneau-etal-2020-unsupervised} for datasets from other domains. 

\paragraph{Training} 
During training, we fine-tune the pretrained contextual embeddings by AdamW \citep{loshchilov2018decoupled} optimizer with a batch size of $4$. We use a learning rate of $5\times 10^{-6}$ to update the parameters in the pretrained contextual embeddings. For the CRF layer parameters, we use a learning rate of $0.05$. We train the NER models for $10$ epochs for the datasets in Social Media and Biomedical domains while we train the NER models for $5$ epochs for other datasets for efficiency as these datasets have more training sentences.

\begin{table*}[ht!]
\centering
\setlength\tabcolsep{5pt}
\small
\begin{tabular}{l||cc|cc|cc|c}
\hlineB{4}
& \multicolumn{2}{c|}{\textbf{Social Media}} & \multicolumn{2}{c|}{\textbf{News}} & \multicolumn{2}{c|}{\textbf{Biomedical}} & \multirow{2}{*}{\textbf{E-commerce}}\\
& WNUT-16 & WNUT-17 & CoNLL-03 & CoNLL++ & BC5CDR & NCBI & \\
\hline\hline
\citet{zhou-etal-2019-dual} & 55.43 & 42.83 & - & - & - & - & - \\
\citet{nguyen-etal-2020-bertweet} & 52.10 & 56.50 & - & - & - & - & - \\
\citet{nie-etal-2020-named} & 55.01 & 50.36  & - & - & - & - & -  \\
\hline
\citet{baevski-etal-2019-cloze} & - & - & 93.50 & - & - & - & -  \\
\citet{wang-etal-2019-crossweigh} & - & - & 93.43 & 94.28 & - & - & -  \\
\citet{li-etal-2020-dice} & - & - & 93.33 & - & - & - & -  \\
\hline
\citet{nooralahzadeh-etal-2019-reinforcement} & - & - & - & - & 89.93 & - & - \\
Bio-Flair \shortcite{sharma2019bioflair} & - & - & - & - & 89.42 & 88.85 & - \\
Bio-BERT \shortcite{lee2020biobert} & - & - & - & - & - & 87.70 & - \\
\hline\hline
  \multicolumn{8}{c}{Evaluation: {\sc \textbf{w/o Context}}}\\
\hline
{\bf\textsc{LUKE}} \shortcite{yamada-etal-2020-luke} & 54.04 & 55.22 & 92.42 & 93.99 & 89.18 & 87.62 & 77.64 \\
{\bf\textsc{w/o Context}} & 56.04 & 57.86 & 93.03 & 94.20 & 90.52 & 88.65 & 81.47\\
{\bf\textsc{CL-$L_2$}} & 57.35\rlap{$^{\dagger}$} & 58.68\rlap{$^{\dagger}$} & 93.08  & 94.38\rlap{$^{\dagger}$} & 90.70\rlap{$^{\dagger}$} & 89.20\rlap{$^{\dagger}$} & 82.43\rlap{$^{\dagger}$}\\
{\bf\textsc{CL-KL}} & 58.14\rlap{$^{\dagger}$} & 59.33\rlap{$^{\dagger}$} & 93.21\rlap{$^{\dagger}$}  & 94.55\rlap{$^{\dagger}$} & 90.73\rlap{$^{\dagger}$} & \textbf{89.24}\rlap{$^{\dagger}$} & 82.31\rlap{$^{\dagger}$}\\
\hline
  \multicolumn{8}{c}{Evaluation: {\sc \textbf{w/ Context}}}\\
\hline
{\bf\textsc{w/ Context}} & 57.43\rlap{$^{\dagger}$} & 60.20\rlap{$^{\dagger}$} & 93.27\rlap{$^{\dagger}$} & 94.56\rlap{$^{\dagger}$} & 90.76\rlap{$^{\dagger}$} & 89.01\rlap{$^{\dagger}$} & 83.15\rlap{$^{\dagger}$}\\
{\bf\textsc{CL-$L_2$}} & 58.61\rlap{$^{\dagger}$} & 60.26\rlap{$^{\dagger}$} & 93.47\rlap{$^{\dagger}$} & 94.62\rlap{$^{\dagger}$} & \textbf{90.99}\rlap{$^{\dagger}$} & 89.22\rlap{$^{\dagger}$} & 83.87\rlap{$^{\dagger}$}\\
{\bf\textsc{CL-KL}} & \textbf{58.98}\rlap{$^{\dagger}$} & \textbf{60.45}\rlap{$^{\dagger}$} & \textbf{93.56}\rlap{$^{\dagger}$} & \textbf{94.81}\rlap{$^{\dagger}$} & 90.93\rlap{$^{\dagger}$} & 88.96\rlap{$^{\dagger}$} & \textbf{83.99}\rlap{$^{\dagger}$}\\
\hlineB{4}
\end{tabular}
\caption{A comparison among recent state-of-the-art models, the baseline and our approaches. ${\dagger}$ represents the model is significantly stronger than the baseline model ({\bf\textsc{w/o Context}}) with $p<0.05$ on Student's T test.}
\label{tab:main}
\end{table*}

\subsection{Results}
We experiment on the following approaches:
\begin{itemize}[leftmargin=*]
    \item {\sc\textbf{LUKE}} is a very recent state-of-the art model on CoNLL-03 NER dataset proposed by \citet{yamada-etal-2020-luke}. We use the same parameter setting as \citet{yamada-etal-2020-luke} and use a single sentence as the input instead of taking document-level contexts in the dataset as in \citet{yamada-etal-2020-luke} for fair comparison.
    \item {\sc\textbf{w/o Context}} represents training the NER model without external contexts (Eq. \ref{eq:nll_loss}), which is the baseline of our approaches. 
    \item {\sc\textbf{w/ Context}} represents training the NER model with external contexts (Eq. \ref{eq:nll_loss_x}). 
    \item {\sc\textbf{CL-$L_2$}} represents minimizing the $L_2$ distance between token representations (Eq. \ref{eq:L2_loss}).
    \item {\sc\textbf{CL-KL}} represents minimizing the KL divergence (Eq. \ref{eq:cl_posterior_loss}) between CRF output distributions.
\end{itemize}
Besides, we also compare our approaches with previous state-of-the-art approaches over entity-level F1 scores\footnote{We do not compare the results from previous work such as \citet{yu-etal-2020-named,luoma-pyysalo-2020-exploring,yamada-etal-2020-luke} that utilizes the document-level contexts in CoNLL-03 NER here. We conduct a comparison with these approaches in Appendix \ref{app:versus}. }.
During the evaluation, our approaches are evaluated using inputs without external contexts ({\sc\textbf{w/o Context}}) and inputs with them ({\sc\textbf{w/ Context}}). We report the results averaged over 5 runs in our experiments. The results are listed in Table \ref{tab:main}\footnote{For the result of Bio-BERT \citep{lee2020biobert} on NCBI-disease dataset, we report the results reported in official code (\url{https://github.com/dmis-lab/biobert}). The results (89.71 in NCBI-disease) reported in the paper used token-level F1 score instead of entity-level F1 score.}. With the external contexts, our models with CL outperform previous state-of-the-art approaches on most of the datasets. Our approaches significantly outperform the baseline that is trained without external contexts. 
Comparing with LUKE, our approaches and our baseline outperform LUKE in all the cases. The possible reason is that LUKE is pretrained only using long word sequences, which makes the model prone to fail to capture the information of entities based on short sentences\footnote{We have confirmed with the authors of LUKE \citep{yamada-etal-2020-luke} that the accuracy on the CoNLL-03 dataset is consistent with their experimental results.}. For our approaches, with CL, the accuracy can be improved on both input views comparing with {\sc \textbf{w/o Context}} and {\sc \textbf{w/ Context}}, which shows adding constraints between the two views during training helps the model better utilize the original text information. For the two constraints in CL, we find that \textbf{CL-KL} is relatively stronger than \textbf{CL-$L_2$} in a majority of the cases. 

\begin{table}[t!]
\centering
\setlength\tabcolsep{5pt}
\small
\begin{tabular}{l|cc}
\hlineB{4}
 & \multicolumn{2}{c}{Evaluation} \\
 & \multicolumn{2}{c}{\textbf{Science and Technology}} \\
 \hline
Approach & {\sc \textbf{w/o Context}} & {\sc \textbf{w/ Context}} \\
\hline\hline
\citet{jia-etal-2019-cross} & 73.59 & - \\
\hline
{\sc\textbf{w/o Context}} & 75.87 & 75.74  \\
{\sc\textbf{w/ Context }} & 75.72 & 75.94  \\
{\sc\textbf{CL-$L_2$ }} & 76.16 & 76.10  \\
{\sc\textbf{CL-KL }} & \textbf{76.37} & \textbf{76.38}  \\
\hlineB{4}
\end{tabular}
\caption{A comparison of different approaches in transfer learning. The models are trained on the CoNLL-03 dataset.}
\label{tab:transfer}
\end{table}

\subsection{Cross-Domain Transfer}
For cross-domain transfer, we train the models on the CoNLL-03 datasets, evaluate the accuracy on the CBS SciTech News dataset, and compare the results with those in \citet{jia-etal-2019-cross}. We evaluate our approaches with each input view and the results are shown in Table \ref{tab:transfer}. Our approaches can improve the accuracy in cross-domain evaluation. The external contexts during evaluation can help to improve the accuracy of {\sc\textbf{w/ Context}}. However, the gap between the two input views for the CL approaches is diminished. The observation shows that CL is able to improve the accuracy in cross-domain transfer for both views and eliminate the gap between the two views.

\begin{table}[t!]
\centering
\setlength\tabcolsep{5pt}
\small
\begin{tabular}{l|cc}
\hlineB{4}
 & \multicolumn{2}{c}{Evaluation} \\
 \hline
Approach & {\sc \textbf{w/o Context}} & {\sc \textbf{w/ Context}} \\
\hline\hline
{\sc\textbf{CL-$L_2$}} & 82.43 & 83.87  \\
{\sc\textbf{CL-KL}} & 82.31 & 83.99  \\
\hline
{\sc\textbf{CL--$L_2$+Semi}} & \textbf{82.88}\rlap{$^{\dagger}$} & 83.92  \\
{\sc\textbf{CL-KL+Semi}} & 82.58\rlap{$^{\dagger}$} & \textbf{84.10}  \\
\hlineB{4}
\end{tabular}
\caption{A comparison between of CL approaches with and without semi-supervised learning. {\sc \textbf{Semi}} represents the approaches with semi-supervised learning. ${\dagger}$ represents the approach is significantly ($p<0.05$) stronger than the approach without semi-supervised learning with the same input view.}
\label{tab:semi}
\end{table}

\subsection{Semi-supervised Cooperative Learning}
Cooperative learning can take advantage of large amounts of unlabeled text for further improvement. We jointly train on the labeled data and unlabeled data in training to form a semi-supervised training manner. During training, we alternate between minimizing the loss (Eq. \ref{eq:final_loss}) for labeled data and the CL loss for unlabeled data (Eq. \ref{eq:cl_loss}). We conduct the experiment on the E-commerce dataset as an example. Results in Table \ref{tab:semi} show that the accuracy of both input views can be improved especially for the input without external contexts, which shows the effectiveness of CL in semi-supervised learning.

\section{Analysis}
We use the WNUT-17 dataset in the analysis.
\subsection{Comparison of Re-ranking Approaches}
Various re-ranking approaches may affect the token representations of the model. We compare our approach with three other re-ranking approaches. The first is the ranking from the search engine without any re-ranking approaches. The second is re-ranking through a fuzzy match score. The approach has been widely applied in a lot of previous work \citep{gu2018search,zhang-etal-2018-guiding,hayati-etal-2018-retrieval,xu-etal-2020-boosting}. The third is BERTScore with tf-idf importance weighting which makes rare words more indicative than common words in scoring. We train our models ({\sc \textbf{w/ Context}}) with external contexts from these re-ranking approaches and report the averaged and best results on WNUT-17 in Table \ref{tab:ranking}. Our results show that re-ranking with BERTScore performs the best, which shows the semantic relevance is helpful for the performance. However, for BERTScore with the tf-idf weighting, the accuracy of the model drops significantly (with $p<0.05$). The possible reason might be that the tf-idf weighting gives high weights to irrelevant texts with rare words during re-ranking.

\begin{table}[t]
\centering
\small
\begin{tabular}{l|cccc}
\hlineB{4}
& \textbf{SE} & \textbf{FM} & \textbf{BS} & \textbf{BS+tf-idf}\\
\hline
\textbf{\textsc{Avg.}} & 59.95 & 59.54 & \textbf{60.20} & 59.71 \\ 
\hline
\textbf{\textsc{Best}} & 61.79 & 60.89 & \textbf{62.29} & 60.96 \\ 
\hlineB{4}
\end{tabular}
\caption{A comparison of different re-ranking approaches by the F1 scores on WNUT-17. \textbf{SE}: Search engine. \textbf{FM}: Fuzzy match score. \textbf{BS}: BERTScore.}
\label{tab:ranking}
\end{table}

\begin{table}[t]
\centering
\setlength\tabcolsep{5pt}
\small
\begin{tabular}{l|c}
\hlineB{4}
& WNUT-17 \\
\hline
\textbf{w/ Context (Ours)} & \textbf{60.20} \\
\hline
\textbf{w/o Context } & 57.86 \\
\textbf{w/ Context (Dataset)} & 57.21\\
\textbf{w/ Context (Generated)} & 57.71\\
\textbf{w/ Context (Random Retrieved)} & 57.53 \\
\textbf{w/ Context (Random Data)} & 47.69\\
\hlineB{4}
\end{tabular}
\caption{A comparison among different contexts types.}
\label{tab:random}
\end{table}

\subsection{How the Context Quality Affects Accuracy}
We analyze how the NER model will perform when the quality of external contexts varies. We train and evaluate the NER model in four conditions with various contexts. The first one takes each dataset split as a document and encodes each sentence with document-level contexts. In this case, we encode the document-level contexts following the approach of \citet{yamada-etal-2020-luke}. The second one uses GPT-2 \citep{radford2019language} to generate 6 relevant sentences as external contexts. The other two conditions randomly select from the retrieved texts or the dataset as external contexts. Results in Table \ref{tab:random} show that all these conditions result in inferior accuracy comparing with the model without any external context. However, our external contexts are more semantically relevant to the input sentence and helpful for prediction.

\begin{table}[t]
\centering
\setlength\tabcolsep{5pt}
\small
\begin{tabular}{l|cc}
\hlineB{4}
 & \multicolumn{2}{c}{Evaluation} \\
 \hline
Approach & {\sc \textbf{w/o Context}} & {\sc \textbf{w/ Context}} \\
\hline\hline
{\sc\textbf{w/o Context }} & 57.86 & 59.40  \\
{\sc\textbf{w/ Context }} & 57.46 & 60.20  \\
{\sc\textbf{w/o CL}} & 58.14 & 59.64 \\
{\sc\textbf{CL-$L_2$ + CL-KL}} & 58.69 & 60.16 \\
\hline
{\sc\textbf{CL-$L_2$ }} & 58.68 & 60.26  \\
{\sc\textbf{CL-KL }} & \textbf{59.33} & \textbf{60.45}  \\
\hlineB{4}
\end{tabular}
\caption{An ablation study of the training and prediction of models.}
\label{tab:ablation}
\end{table}

\subsection{Ablation Study}
To show the effectiveness of CL, we conduct three ablation studies for our approach. The first one is training the NER model based on one view and predict on the other. The second is jointly training both views without the CL loss term (removing $\mcL_{\text{CL}}(\theta)$ in Eq. \ref{eq:final_loss}). The final one is using both CL losses to train the model ($\mcL_{\text{CL}}(\theta)=\mcL_{\text{CL-$L_2$}}(\theta)+\mcL_{\text{CL-KL}}(\theta)$ in Eq. \ref{eq:final_loss}). Results in Table \ref{tab:ablation} show that the external context can help to improve the accuracy even when the NER model is trained without the contexts. However, when the model is trained with the external contexts, the accuracy of the model drops when predicting the inputs without external contexts. In joint training without CL, the accuracy of the model over inputs without contexts can be slightly improved but the accuracy over inputs with contexts drops, which shows the benefit of adding CL. For the model trained with both CL losses, we find no improvement over the models trained with a single CL loss. 

\section{Related Work}
\paragraph{Named Entity Recognition}
Named Entity Recognition \citep{Sundheim1995NamedET} has been studied for decades. Most of the work takes NER as a sequence labeling problem and applies the linear-chain CRF \citep{10.5555/645530.655813} to achieve state-of-the-art accuracy \citep{ma-hovy-2016-end,lample-etal-2016-neural,akbik-etal-2018-contextual,akbik-etal-2019-pooled,wang-etal-2020-more}. 
Recently, the improvement of accuracy mainly benefits from stronger token representations such as pretrained contextual embeddings such as BERT \citep{devlin-etal-2019-bert}, Flair \citep{akbik-etal-2018-contextual} and LUKE \citep{yamada-etal-2020-luke}. Very recent work \citep{yu-etal-2020-named,yamada-etal-2020-luke} utilizes the strength of pretrained contextual embeddings over long-range dependency and encodes the document-level contexts for token representations to achieve state-of-the-art accuracy on CoNLL 2002/2003 NER datasets \citep{tjong-kim-sang-2002-introduction,tjong-kim-sang-de-meulder-2003-introduction}. 

\paragraph{Improving Models through Retrieval}
Retrieving related texts from a certain database (such as the training set) has been widely applied in tasks such as neural machine translation \citep{gu2018search,zhang-etal-2018-guiding,xu-etal-2020-boosting}, text generation \citep{weston-etal-2018-retrieve,kim-etal-2020-retrieval}, semantic parsing \citep{hashimoto2018retrieve,guo-etal-2019-coupling}. Most of the work uses the retrieved texts to guide the generation or refine the retrieved texts through the neural model, while we take the retrieved texts as the contexts of the input sentence to improve the semantic representations of the input tokens.
For the re-ranking models, fuzzy match score \citep{gu2018search,zhang-etal-2018-guiding,hayati-etal-2018-retrieval,xu-etal-2020-boosting}, attention mechanisms \citep{cao-etal-2018-retrieve,cai-etal-2019-retrieval}, and dot products between sentence representations \citep{lewis2020retrieval,xu-etal-2020-boosting} are usual scoring functions to re-rank the retrieved texts. Instead, we use BERTScore to re-rank the retrieved texts instead as BERTScore evaluates semantic correlations between the texts based on pretrained contextual embeddings.

\paragraph{Multi-View Learning}
Multi-View Learning is a technique applied to inputs that can be split into multiple subsets. Co-training \citep{blum1998combining} and co-regularization \citep{sindhwani2005co} train a separate model for each view. These approaches are semi-supervised learning techniques that require two independent views of the data. The model with higher confidence is applied to construct additional labeled data by predicting on unlabeled data. \citet{sun2013survey} and \citet{xu2013survey} have extensively studied various multi-view learning approaches. \citet{hu2021multi} shows the effectiveness of multi-view learning on cross-lingual structured prediction tasks. Recently, \citet{clark-etal-2018-semi} proposed Cross-View Training (CVT), which trains a unified model instead of multiple models and targets at minimizing the KL divergence between the probability distributions of the model and auxiliary prediction modules. Comparing with CVT, CL targets at improving the accuracy of two kinds of inputs rather than only one of them. We also propose to minimize the distance of token representations between different views in addition to KL-divergence. Besides, CL utilizes the external contexts and therefore we do not need to construct auxiliary prediction modules in the model. Moreover, CVT cannot be directly applied to our transformer-based embeddings. Finally, our decoding layer in the model uses the CRF layer instead of the simple Softmax layer as in CVT. The CRF layer is stronger but more difficult for KL-divergence computation.

\paragraph{Knowledge Distillation}
Knowledge distillation \citep{Bucilua:2006:MC:1150402.1150464,44873} transfers the knowledge of ``teacher'' models to smaller ``student'' models through minimizing the KL divergence of prediction probability distribution between the models. In speech recognition \citep{Huang2018} and natural language processing \citep{wang-etal-2020-structure,wang2020structural}, the marginal probability distribution of the linear-chain CRF layer has been applied to distill the knowledge between teacher models and student models. Comparing with these approaches, our approaches train a single unified model instead of transferring the knowledge between two models. We also show that the accuracy of both views can be improved with our approaches, unlike in knowledge distillation only the student model is updated and improved.

\section{Conclusion}
In this paper, we propose to improve the NER model's accuracy by retrieving related contexts from a search engine as external contexts of the inputs. To improve the robustness of the models when no external contexts are available, we propose Cooperative Learning. Cooperative Learning adds constraints between two input views over either the token representations or label distributions of both input views to be consistent. Empirical results show that our approach significantly outperforms the baseline models and previous state-of-the-art approaches on the datasets over 5 domains. We also show the effectiveness of Cooperative Learning in a semi-supervised training manner. 

\section*{Acknowledgments}
This work was supported by the National Natural Science Foundation of China (61976139) and by Alibaba Group through Alibaba Innovative Research Program. We thank Kaibo Zhang for his help in crawling related texts from Google Search and thank Jiong Cai and Zhuo Chen for their comments and suggestions on writing. 


\bibliographystyle{acl_natbib}
\bibliography{anthology,acl2021}

\begin{thebibliography}{56}
\expandafter\ifx\csname natexlab\endcsname\relax\def\natexlab#1{#1}\fi

\bibitem[{Akbik et~al.(2019)Akbik, Bergmann, and
  Vollgraf}]{akbik-etal-2019-pooled}
Alan Akbik, Tanja Bergmann, and Roland Vollgraf. 2019.
\newblock \href {https://doi.org/10.18653/v1/N19-1078} {Pooled contextualized
  embeddings for named entity recognition}.
\newblock In \emph{Proceedings of the 2019 Conference of the North {A}merican
  Chapter of the Association for Computational Linguistics: Human Language
  Technologies, Volume 1 (Long and Short Papers)}, pages 724--728, Minneapolis,
  Minnesota. Association for Computational Linguistics.

\bibitem[{Akbik et~al.(2018)Akbik, Blythe, and
  Vollgraf}]{akbik-etal-2018-contextual}
Alan Akbik, Duncan Blythe, and Roland Vollgraf. 2018.
\newblock \href {https://www.aclweb.org/anthology/C18-1139} {Contextual string
  embeddings for sequence labeling}.
\newblock In \emph{Proceedings of the 27th International Conference on
  Computational Linguistics}, pages 1638--1649, Santa Fe, New Mexico, USA.
  Association for Computational Linguistics.

\bibitem[{Baevski et~al.(2019)Baevski, Edunov, Liu, Zettlemoyer, and
  Auli}]{baevski-etal-2019-cloze}
Alexei Baevski, Sergey Edunov, Yinhan Liu, Luke Zettlemoyer, and Michael Auli.
  2019.
\newblock \href {https://doi.org/10.18653/v1/D19-1539} {Cloze-driven
  pretraining of self-attention networks}.
\newblock In \emph{Proceedings of the 2019 Conference on Empirical Methods in
  Natural Language Processing and the 9th International Joint Conference on
  Natural Language Processing (EMNLP-IJCNLP)}, pages 5360--5369, Hong Kong,
  China. Association for Computational Linguistics.

\bibitem[{Blum and Mitchell(1998)}]{blum1998combining}
Avrim Blum and Tom Mitchell. 1998.
\newblock Combining labeled and unlabeled data with co-training.
\newblock In \emph{Proceedings of the eleventh annual conference on
  Computational learning theory}, pages 92--100.

\bibitem[{Bucilu\v{a} et~al.(2006)Bucilu\v{a}, Caruana, and
  Niculescu-Mizil}]{Bucilua:2006:MC:1150402.1150464}
Cristian Bucilu\v{a}, Rich Caruana, and Alexandru Niculescu-Mizil. 2006.
\newblock \href {https://doi.org/10.1145/1150402.1150464} {Model compression}.
\newblock In \emph{Proceedings of the 12th ACM SIGKDD International Conference
  on Knowledge Discovery and Data Mining}, KDD '06, pages 535--541, New York,
  NY, USA. ACM.

\bibitem[{Cai et~al.(2019)Cai, Wang, Bi, Tu, Liu, and
  Shi}]{cai-etal-2019-retrieval}
Deng Cai, Yan Wang, Wei Bi, Zhaopeng Tu, Xiaojiang Liu, and Shuming Shi. 2019.
\newblock \href {https://doi.org/10.18653/v1/D19-1195} {Retrieval-guided
  dialogue response generation via a matching-to-generation framework}.
\newblock In \emph{Proceedings of the 2019 Conference on Empirical Methods in
  Natural Language Processing and the 9th International Joint Conference on
  Natural Language Processing (EMNLP-IJCNLP)}, pages 1866--1875, Hong Kong,
  China. Association for Computational Linguistics.

\bibitem[{Cao et~al.(2018)Cao, Li, Li, and Wei}]{cao-etal-2018-retrieve}
Ziqiang Cao, Wenjie Li, Sujian Li, and Furu Wei. 2018.
\newblock \href {https://doi.org/10.18653/v1/P18-1015} {Retrieve, rerank and
  rewrite: Soft template based neural summarization}.
\newblock In \emph{Proceedings of the 56th Annual Meeting of the Association
  for Computational Linguistics (Volume 1: Long Papers)}, pages 152--161,
  Melbourne, Australia. Association for Computational Linguistics.

\bibitem[{Clark et~al.(2018)Clark, Luong, Manning, and
  Le}]{clark-etal-2018-semi}
Kevin Clark, Minh-Thang Luong, Christopher~D. Manning, and Quoc Le. 2018.
\newblock \href {https://doi.org/10.18653/v1/D18-1217} {Semi-supervised
  sequence modeling with cross-view training}.
\newblock In \emph{Proceedings of the 2018 Conference on Empirical Methods in
  Natural Language Processing}, pages 1914--1925, Brussels, Belgium.
  Association for Computational Linguistics.

\bibitem[{Conneau et~al.(2020)Conneau, Khandelwal, Goyal, Chaudhary, Wenzek,
  Guzm{\'a}n, Grave, Ott, Zettlemoyer, and
  Stoyanov}]{conneau-etal-2020-unsupervised}
Alexis Conneau, Kartikay Khandelwal, Naman Goyal, Vishrav Chaudhary, Guillaume
  Wenzek, Francisco Guzm{\'a}n, Edouard Grave, Myle Ott, Luke Zettlemoyer, and
  Veselin Stoyanov. 2020.
\newblock \href {https://doi.org/10.18653/v1/2020.acl-main.747} {Unsupervised
  cross-lingual representation learning at scale}.
\newblock In \emph{Proceedings of the 58th Annual Meeting of the Association
  for Computational Linguistics}, pages 8440--8451, Online. Association for
  Computational Linguistics.

\bibitem[{Derczynski et~al.(2017)Derczynski, Nichols, van Erp, and
  Limsopatham}]{derczynski-etal-2017-results}
Leon Derczynski, Eric Nichols, Marieke van Erp, and Nut Limsopatham. 2017.
\newblock \href {https://doi.org/10.18653/v1/W17-4418} {Results of the
  {WNUT}2017 shared task on novel and emerging entity recognition}.
\newblock In \emph{Proceedings of the 3rd Workshop on Noisy User-generated
  Text}, pages 140--147, Copenhagen, Denmark. Association for Computational
  Linguistics.

\bibitem[{Devlin et~al.(2019)Devlin, Chang, Lee, and
  Toutanova}]{devlin-etal-2019-bert}
Jacob Devlin, Ming-Wei Chang, Kenton Lee, and Kristina Toutanova. 2019.
\newblock \href {https://doi.org/10.18653/v1/N19-1423} {{BERT}: Pre-training of
  deep bidirectional transformers for language understanding}.
\newblock In \emph{Proceedings of the 2019 Conference of the North {A}merican
  Chapter of the Association for Computational Linguistics: Human Language
  Technologies, Volume 1 (Long and Short Papers)}, pages 4171--4186,
  Minneapolis, Minnesota. Association for Computational Linguistics.

\bibitem[{Do{\u{g}}an et~al.(2014)Do{\u{g}}an, Leaman, and Lu}]{dougan2014ncbi}
Rezarta~Islamaj Do{\u{g}}an, Robert Leaman, and Zhiyong Lu. 2014.
\newblock Ncbi disease corpus: a resource for disease name recognition and
  concept normalization.
\newblock \emph{Journal of biomedical informatics}, 47:1--10.

\bibitem[{Gu et~al.(2018)Gu, Wang, Cho, and Li}]{gu2018search}
Jiatao Gu, Yong Wang, Kyunghyun Cho, and Victor~OK Li. 2018.
\newblock Search engine guided neural machine translation.
\newblock In \emph{Proceedings of the AAAI Conference on Artificial
  Intelligence}, volume~32.

\bibitem[{Guo et~al.(2019)Guo, Tang, Duan, Zhou, and
  Yin}]{guo-etal-2019-coupling}
Daya Guo, Duyu Tang, Nan Duan, Ming Zhou, and Jian Yin. 2019.
\newblock \href {https://doi.org/10.18653/v1/P19-1082} {Coupling retrieval and
  meta-learning for context-dependent semantic parsing}.
\newblock In \emph{Proceedings of the 57th Annual Meeting of the Association
  for Computational Linguistics}, pages 855--866, Florence, Italy. Association
  for Computational Linguistics.

\bibitem[{Hashimoto et~al.(2018)Hashimoto, Guu, Oren, and
  Liang}]{hashimoto2018retrieve}
Tatsunori~B Hashimoto, Kelvin Guu, Yonatan Oren, and Percy Liang. 2018.
\newblock A retrieve-and-edit framework for predicting structured outputs.
\newblock In \emph{Proceedings of the 32nd International Conference on Neural
  Information Processing Systems}, pages 10073--10083.

\bibitem[{Hayati et~al.(2018)Hayati, Olivier, Avvaru, Yin, Tomasic, and
  Neubig}]{hayati-etal-2018-retrieval}
Shirley~Anugrah Hayati, Raphael Olivier, Pravalika Avvaru, Pengcheng Yin,
  Anthony Tomasic, and Graham Neubig. 2018.
\newblock \href {https://doi.org/10.18653/v1/D18-1111} {Retrieval-based neural
  code generation}.
\newblock In \emph{Proceedings of the 2018 Conference on Empirical Methods in
  Natural Language Processing}, pages 925--930, Brussels, Belgium. Association
  for Computational Linguistics.

\bibitem[{Hinton et~al.(2015)Hinton, Vinyals, and Dean}]{44873}
Geoffrey Hinton, Oriol Vinyals, and Jeffrey Dean. 2015.
\newblock \href {http://arxiv.org/abs/1503.02531} {Distilling the knowledge in
  a neural network}.
\newblock In \emph{NIPS Deep Learning and Representation Learning Workshop}.

\bibitem[{Hu et~al.(2021)Hu, Jiang, Bach, Wang, Huang, Huang, and
  Tu}]{hu2021multi}
zechuan Hu, Yong Jiang, Nguyen Bach, Tao Wang, Zhongqiang Huang, Fei Huang, and
  Kewei Tu. 2021.
\newblock {{Multi-View Cross-Lingual Structured Prediction with Minimum
  Supervision}}.
\newblock In \emph{{the Joint Conference of the 59th Annual Meeting of the
  Association for Computational Linguistics and the 11th International Joint
  Conference on Natural Language Processing (\textbf{ACL-IJCNLP 2021})}}.
  Association for Computational Linguistics.

\bibitem[{Huang et~al.(2018)Huang, You, Chen, Qian, and Yu}]{Huang2018}
Mingkun Huang, Yongbin You, Zhehuai Chen, Yanmin Qian, and Kai Yu. 2018.
\newblock \href {https://doi.org/10.21437/Interspeech.2018-1589} {Knowledge
  distillation for sequence model}.
\newblock In \emph{Proc. Interspeech 2018}, pages 3703--3707.

\bibitem[{Jia et~al.(2019)Jia, Liang, and Zhang}]{jia-etal-2019-cross}
Chen Jia, Xiaobo Liang, and Yue Zhang. 2019.
\newblock \href {https://doi.org/10.18653/v1/P19-1236} {Cross-domain {NER}
  using cross-domain language modeling}.
\newblock In \emph{Proceedings of the 57th Annual Meeting of the Association
  for Computational Linguistics}, pages 2464--2474, Florence, Italy.
  Association for Computational Linguistics.

\bibitem[{Kim et~al.(2020)Kim, Choi, Amplayo, and
  Hwang}]{kim-etal-2020-retrieval}
Jihyeok Kim, Seungtaek Choi, Reinald~Kim Amplayo, and Seung-won Hwang. 2020.
\newblock \href {https://www.aclweb.org/anthology/2020.coling-main.207}
  {Retrieval-augmented controllable review generation}.
\newblock In \emph{Proceedings of the 28th International Conference on
  Computational Linguistics}, pages 2284--2295, Barcelona, Spain (Online).
  International Committee on Computational Linguistics.

\bibitem[{Lafferty et~al.(2001)Lafferty, McCallum, and
  Pereira}]{10.5555/645530.655813}
John~D. Lafferty, Andrew McCallum, and Fernando C.~N. Pereira. 2001.
\newblock Conditional random fields: Probabilistic models for segmenting and
  labeling sequence data.
\newblock In \emph{Proceedings of the Eighteenth International Conference on
  Machine Learning}, ICML ’01, page 282–289, San Francisco, CA, USA. Morgan
  Kaufmann Publishers Inc.

\bibitem[{Lample et~al.(2016)Lample, Ballesteros, Subramanian, Kawakami, and
  Dyer}]{lample-etal-2016-neural}
Guillaume Lample, Miguel Ballesteros, Sandeep Subramanian, Kazuya Kawakami, and
  Chris Dyer. 2016.
\newblock \href {https://doi.org/10.18653/v1/N16-1030} {Neural architectures
  for named entity recognition}.
\newblock In \emph{Proceedings of the 2016 Conference of the North {A}merican
  Chapter of the Association for Computational Linguistics: Human Language
  Technologies}, pages 260--270, San Diego, California. Association for
  Computational Linguistics.

\bibitem[{Lee et~al.(2020)Lee, Yoon, Kim, Kim, Kim, So, and
  Kang}]{lee2020biobert}
Jinhyuk Lee, Wonjin Yoon, Sungdong Kim, Donghyeon Kim, Sunkyu Kim, Chan~Ho So,
  and Jaewoo Kang. 2020.
\newblock Biobert: a pre-trained biomedical language representation model for
  biomedical text mining.
\newblock \emph{Bioinformatics}, 36(4):1234--1240.

\bibitem[{Lewis et~al.(2020)Lewis, Perez, Piktus, Petroni, Karpukhin, Goyal,
  K{\"u}ttler, Lewis, Yih, Rockt{\"a}schel et~al.}]{lewis2020retrieval}
Patrick Lewis, Ethan Perez, Aleksandara Piktus, Fabio Petroni, Vladimir
  Karpukhin, Naman Goyal, Heinrich K{\"u}ttler, Mike Lewis, Wen-tau Yih, Tim
  Rockt{\"a}schel, et~al. 2020.
\newblock Retrieval-augmented generation for knowledge-intensive nlp tasks.
\newblock \emph{arXiv preprint arXiv:2005.11401}.

\bibitem[{Li et~al.(2016)Li, Sun, Johnson, Sciaky, Wei, Leaman, Davis,
  Mattingly, Wiegers, and Lu}]{li2016biocreative}
Jiao Li, Yueping Sun, Robin~J Johnson, Daniela Sciaky, Chih-Hsuan Wei, Robert
  Leaman, Allan~Peter Davis, Carolyn~J Mattingly, Thomas~C Wiegers, and Zhiyong
  Lu. 2016.
\newblock Biocreative v cdr task corpus: a resource for chemical disease
  relation extraction.
\newblock \emph{Database: The Journal of Biological Databases and Curation},
  2016.

\bibitem[{Li et~al.(2020)Li, Sun, Meng, Liang, Wu, and Li}]{li-etal-2020-dice}
Xiaoya Li, Xiaofei Sun, Yuxian Meng, Junjun Liang, Fei Wu, and Jiwei Li. 2020.
\newblock \href {https://doi.org/10.18653/v1/2020.acl-main.45} {Dice loss for
  data-imbalanced {NLP} tasks}.
\newblock In \emph{Proceedings of the 58th Annual Meeting of the Association
  for Computational Linguistics}, pages 465--476, Online. Association for
  Computational Linguistics.

\bibitem[{Liu et~al.(2019)Liu, Ott, Goyal, Du, Joshi, Chen, Levy, Lewis,
  Zettlemoyer, and Stoyanov}]{liu2019roberta}
Yinhan Liu, Myle Ott, Naman Goyal, Jingfei Du, Mandar Joshi, Danqi Chen, Omer
  Levy, Mike Lewis, Luke Zettlemoyer, and Veselin Stoyanov. 2019.
\newblock Roberta: A robustly optimized bert pretraining approach.
\newblock \emph{arXiv preprint arXiv:1907.11692}.

\bibitem[{Loshchilov and Hutter(2018)}]{loshchilov2018decoupled}
Ilya Loshchilov and Frank Hutter. 2018.
\newblock Decoupled weight decay regularization.
\newblock In \emph{International Conference on Learning Representations}.

\bibitem[{Luoma and Pyysalo(2020)}]{luoma-pyysalo-2020-exploring}
Jouni Luoma and Sampo Pyysalo. 2020.
\newblock \href {https://www.aclweb.org/anthology/2020.coling-main.78}
  {Exploring cross-sentence contexts for named entity recognition with {BERT}}.
\newblock In \emph{Proceedings of the 28th International Conference on
  Computational Linguistics}, pages 904--914, Barcelona, Spain (Online).
  International Committee on Computational Linguistics.

\bibitem[{Ma and Hovy(2016)}]{ma-hovy-2016-end}
Xuezhe Ma and Eduard Hovy. 2016.
\newblock \href {https://doi.org/10.18653/v1/P16-1101} {End-to-end sequence
  labeling via bi-directional {LSTM}-{CNN}s-{CRF}}.
\newblock In \emph{Proceedings of the 54th Annual Meeting of the Association
  for Computational Linguistics (Volume 1: Long Papers)}, pages 1064--1074,
  Berlin, Germany. Association for Computational Linguistics.

\bibitem[{Nguyen et~al.(2020)Nguyen, Vu, and
  Tuan~Nguyen}]{nguyen-etal-2020-bertweet}
Dat~Quoc Nguyen, Thanh Vu, and Anh Tuan~Nguyen. 2020.
\newblock \href {https://doi.org/10.18653/v1/2020.emnlp-demos.2} {{BERT}weet: A
  pre-trained language model for {E}nglish tweets}.
\newblock In \emph{Proceedings of the 2020 Conference on Empirical Methods in
  Natural Language Processing: System Demonstrations}, pages 9--14, Online.
  Association for Computational Linguistics.

\bibitem[{Nie et~al.(2020)Nie, Tian, Wan, Song, and Dai}]{nie-etal-2020-named}
Yuyang Nie, Yuanhe Tian, Xiang Wan, Yan Song, and Bo~Dai. 2020.
\newblock \href {https://doi.org/10.18653/v1/2020.emnlp-main.107} {Named entity
  recognition for social media texts with semantic augmentation}.
\newblock In \emph{Proceedings of the 2020 Conference on Empirical Methods in
  Natural Language Processing (EMNLP)}, pages 1383--1391, Online. Association
  for Computational Linguistics.

\bibitem[{Nooralahzadeh et~al.(2019)Nooralahzadeh, L{\o}nning, and
  {\O}vrelid}]{nooralahzadeh-etal-2019-reinforcement}
Farhad Nooralahzadeh, Jan~Tore L{\o}nning, and Lilja {\O}vrelid. 2019.
\newblock \href {https://doi.org/10.18653/v1/D19-6125} {Reinforcement-based
  denoising of distantly supervised {NER} with partial annotation}.
\newblock In \emph{Proceedings of the 2nd Workshop on Deep Learning Approaches
  for Low-Resource NLP (DeepLo 2019)}, pages 225--233, Hong Kong, China.
  Association for Computational Linguistics.

\bibitem[{Peters et~al.(2018)Peters, Neumann, Iyyer, Gardner, Clark, Lee, and
  Zettlemoyer}]{peters-etal-2018-deep}
Matthew Peters, Mark Neumann, Mohit Iyyer, Matt Gardner, Christopher Clark,
  Kenton Lee, and Luke Zettlemoyer. 2018.
\newblock \href {https://doi.org/10.18653/v1/N18-1202} {Deep contextualized
  word representations}.
\newblock In \emph{Proceedings of the 2018 Conference of the North {A}merican
  Chapter of the Association for Computational Linguistics: Human Language
  Technologies, Volume 1 (Long Papers)}, pages 2227--2237, New Orleans,
  Louisiana. Association for Computational Linguistics.

\bibitem[{Radford et~al.(2019)Radford, Wu, Child, Luan, Amodei, and
  Sutskever}]{radford2019language}
Alec Radford, Jeffrey Wu, Rewon Child, David Luan, Dario Amodei, and Ilya
  Sutskever. 2019.
\newblock Language models are unsupervised multitask learners.
\newblock \emph{OpenAI blog}, 1(8):9.

\bibitem[{Sharma and Daniel~Jr(2019)}]{sharma2019bioflair}
Shreyas Sharma and Ron Daniel~Jr. 2019.
\newblock Bioflair: Pretrained pooled contextualized embeddings for biomedical
  sequence labeling tasks.
\newblock \emph{arXiv preprint arXiv:1908.05760}.

\bibitem[{Sindhwani and Niyogi(2005)}]{sindhwani2005co}
Vikas Sindhwani and Partha Niyogi. 2005.
\newblock A co-regularized approach to semi-supervised learning with multiple
  views.
\newblock In \emph{Proceedings of the ICML Workshop on Learning with Multiple
  Views}. Citeseer.

\bibitem[{Strauss et~al.(2016)Strauss, Toma, Ritter, de~Marneffe, and
  Xu}]{strauss-etal-2016-results}
Benjamin Strauss, Bethany Toma, Alan Ritter, Marie-Catherine de~Marneffe, and
  Wei Xu. 2016.
\newblock \href {https://www.aclweb.org/anthology/W16-3919} {Results of the
  {WNUT}16 named entity recognition shared task}.
\newblock In \emph{Proceedings of the 2nd Workshop on Noisy User-generated Text
  ({WNUT})}, pages 138--144, Osaka, Japan. The COLING 2016 Organizing
  Committee.

\bibitem[{Sun(2013)}]{sun2013survey}
Shiliang Sun. 2013.
\newblock A survey of multi-view machine learning.
\newblock \emph{Neural computing and applications}, 23(7-8):2031--2038.

\bibitem[{Sundheim(1995)}]{Sundheim1995NamedET}
Beth~M. Sundheim. 1995.
\newblock Named entity task definition, version 2.1.
\newblock In \emph{Proceedings of the Sixth Message Understanding Conference},
  pages 319--332.

\bibitem[{Tjong Kim~Sang(2002)}]{tjong-kim-sang-2002-introduction}
Erik~F. Tjong Kim~Sang. 2002.
\newblock \href {https://www.aclweb.org/anthology/W02-2024} {Introduction to
  the {C}o{NLL}-2002 shared task: Language-independent named entity
  recognition}.
\newblock In \emph{{COLING}-02: The 6th Conference on Natural Language Learning
  2002 ({C}o{NLL}-2002)}.

\bibitem[{Tjong Kim~Sang and
  De~Meulder(2003)}]{tjong-kim-sang-de-meulder-2003-introduction}
Erik~F. Tjong Kim~Sang and Fien De~Meulder. 2003.
\newblock \href {https://www.aclweb.org/anthology/W03-0419} {Introduction to
  the {C}o{NLL}-2003 shared task: Language-independent named entity
  recognition}.
\newblock In \emph{Proceedings of the Seventh Conference on Natural Language
  Learning at {HLT}-{NAACL} 2003}, pages 142--147.

\bibitem[{Wang et~al.(2020{\natexlab{a}})Wang, Jiang, Bach, Wang, Huang, and
  Tu}]{wang-etal-2020-structure}
Xinyu Wang, Yong Jiang, Nguyen Bach, Tao Wang, Fei Huang, and Kewei Tu.
  2020{\natexlab{a}}.
\newblock \href {https://doi.org/10.18653/v1/2020.acl-main.304}
  {Structure-level knowledge distillation for multilingual sequence labeling}.
\newblock In \emph{Proceedings of the 58th Annual Meeting of the Association
  for Computational Linguistics}, pages 3317--3330, Online. Association for
  Computational Linguistics.

\bibitem[{Wang et~al.(2021{\natexlab{a}})Wang, Jiang, Bach, Wang, Huang, Huang,
  and Tu}]{wang2020automated}
Xinyu Wang, Yong Jiang, Nguyen Bach, Tao Wang, Zhongqiang Huang, Fei Huang, and
  Kewei Tu. 2021{\natexlab{a}}.
\newblock {{Automated Concatenation of Embeddings for Structured Prediction}}.
\newblock In \emph{{the Joint Conference of the 59th Annual Meeting of the
  Association for Computational Linguistics and the 11th International Joint
  Conference on Natural Language Processing (\textbf{ACL-IJCNLP 2021})}}.
  Association for Computational Linguistics.

\bibitem[{Wang et~al.(2020{\natexlab{b}})Wang, Jiang, Bach, Wang, Zhongqiang,
  Huang, and Tu}]{wang-etal-2020-more}
Xinyu Wang, Yong Jiang, Nguyen Bach, Tao Wang, Huang Zhongqiang, Fei Huang, and
  Kewei Tu. 2020{\natexlab{b}}.
\newblock More embeddings, better sequence labelers?
\newblock In \emph{Findings of EMNLP}, Online.

\bibitem[{Wang et~al.(2021{\natexlab{b}})Wang, Jiang, Yan, Jia, Bach, Wang,
  Huang, Huang, and Tu}]{wang2020structural}
Xinyu Wang, Yong Jiang, Zhaohui Yan, Zixia Jia, Nguyen Bach, Tao Wang,
  Zhongqiang Huang, Fei Huang, and Kewei Tu. 2021{\natexlab{b}}.
\newblock {{Structural Knowledge Distillation: Tractably Distilling Information
  for Structured Predictor}}.
\newblock In \emph{{the Joint Conference of the 59th Annual Meeting of the
  Association for Computational Linguistics and the 11th International Joint
  Conference on Natural Language Processing (\textbf{ACL-IJCNLP 2021})}}.
  Association for Computational Linguistics.

\bibitem[{Wang et~al.(2019)Wang, Shang, Liu, Lu, Liu, and
  Han}]{wang-etal-2019-crossweigh}
Zihan Wang, Jingbo Shang, Liyuan Liu, Lihao Lu, Jiacheng Liu, and Jiawei Han.
  2019.
\newblock \href {https://doi.org/10.18653/v1/D19-1519} {{C}ross{W}eigh:
  Training named entity tagger from imperfect annotations}.
\newblock In \emph{Proceedings of the 2019 Conference on Empirical Methods in
  Natural Language Processing and the 9th International Joint Conference on
  Natural Language Processing (EMNLP-IJCNLP)}, pages 5154--5163, Hong Kong,
  China. Association for Computational Linguistics.

\bibitem[{Weston et~al.(2018)Weston, Dinan, and
  Miller}]{weston-etal-2018-retrieve}
Jason Weston, Emily Dinan, and Alexander Miller. 2018.
\newblock \href {https://doi.org/10.18653/v1/W18-5713} {Retrieve and refine:
  Improved sequence generation models for dialogue}.
\newblock In \emph{Proceedings of the 2018 {EMNLP} Workshop {SCAI}: The 2nd
  International Workshop on Search-Oriented Conversational {AI}}, pages 87--92,
  Brussels, Belgium. Association for Computational Linguistics.

\bibitem[{Xu et~al.(2013)Xu, Tao, and Xu}]{xu2013survey}
Chang Xu, Dacheng Tao, and Chao Xu. 2013.
\newblock A survey on multi-view learning.
\newblock \emph{arXiv preprint arXiv:1304.5634}.

\bibitem[{Xu et~al.(2020)Xu, Crego, and Senellart}]{xu-etal-2020-boosting}
Jitao Xu, Josep Crego, and Jean Senellart. 2020.
\newblock \href {https://doi.org/10.18653/v1/2020.acl-main.144} {Boosting
  neural machine translation with similar translations}.
\newblock In \emph{Proceedings of the 58th Annual Meeting of the Association
  for Computational Linguistics}, pages 1580--1590, Online. Association for
  Computational Linguistics.

\bibitem[{Yamada et~al.(2020)Yamada, Asai, Shindo, Takeda, and
  Matsumoto}]{yamada-etal-2020-luke}
Ikuya Yamada, Akari Asai, Hiroyuki Shindo, Hideaki Takeda, and Yuji Matsumoto.
  2020.
\newblock \href {https://doi.org/10.18653/v1/2020.emnlp-main.523} {{LUKE}: Deep
  contextualized entity representations with entity-aware self-attention}.
\newblock In \emph{Proceedings of the 2020 Conference on Empirical Methods in
  Natural Language Processing (EMNLP)}, pages 6442--6454, Online. Association
  for Computational Linguistics.

\bibitem[{Yu et~al.(2020)Yu, Bohnet, and Poesio}]{yu-etal-2020-named}
Juntao Yu, Bernd Bohnet, and Massimo Poesio. 2020.
\newblock \href {https://doi.org/10.18653/v1/2020.acl-main.577} {Named entity
  recognition as dependency parsing}.
\newblock In \emph{Proceedings of the 58th Annual Meeting of the Association
  for Computational Linguistics}, pages 6470--6476, Online. Association for
  Computational Linguistics.

\bibitem[{Zhang et~al.(2018)Zhang, Utiyama, Sumita, Neubig, and
  Nakamura}]{zhang-etal-2018-guiding}
Jingyi Zhang, Masao Utiyama, Eiichro Sumita, Graham Neubig, and Satoshi
  Nakamura. 2018.
\newblock \href {https://doi.org/10.18653/v1/N18-1120} {Guiding neural machine
  translation with retrieved translation pieces}.
\newblock In \emph{Proceedings of the 2018 Conference of the North {A}merican
  Chapter of the Association for Computational Linguistics: Human Language
  Technologies, Volume 1 (Long Papers)}, pages 1325--1335, New Orleans,
  Louisiana. Association for Computational Linguistics.

\bibitem[{Zhang et~al.(2020)Zhang, Kishore, Wu, Weinberger, and
  Artzi}]{Zhang*2020BERTScore:}
Tianyi Zhang, Varsha Kishore, Felix Wu, Kilian~Q. Weinberger, and Yoav Artzi.
  2020.
\newblock \href {https://openreview.net/forum?id=SkeHuCVFDr} {Bertscore:
  Evaluating text generation with bert}.
\newblock In \emph{International Conference on Learning Representations}.

\bibitem[{Zhou et~al.(2019)Zhou, Zhang, Jin, Zhu, Fang, Goh, and
  Kwok}]{zhou-etal-2019-dual}
Joey~Tianyi Zhou, Hao Zhang, Di~Jin, Hongyuan Zhu, Meng Fang, Rick Siow~Mong
  Goh, and Kenneth Kwok. 2019.
\newblock \href {https://doi.org/10.18653/v1/P19-1336} {Dual adversarial neural
  transfer for low-resource named entity recognition}.
\newblock In \emph{Proceedings of the 57th Annual Meeting of the Association
  for Computational Linguistics}, pages 3461--3471, Florence, Italy.
  Association for Computational Linguistics.

\end{thebibliography}

\newpage

\appendix

\begin{table}[t]
\centering
\small
\begin{tabular}{l|c}
\hlineB{4}
 Approach & CoNLL-03 \\
 \hline
\citet{yu-etal-2020-named}\rlap{$^{\dagger}$} & 93.50\\
\citet{yamada-etal-2020-luke} & 94.30\\
\citet{luoma-pyysalo-2020-exploring}\rlap{$^{\dagger}$} & 93.74\\
\citet{wang2020automated} & \textbf{94.60}\\
{\sc\textbf{w/ Doc Context}} & 94.12 \\
\hline
{\sc\textbf{w/o Context }} & 93.30 \\
{\sc\textbf{w/ Context }} & 93.55 \\
{\sc\textbf{CL-$L_2$ }} & 93.68  \\
{\sc\textbf{CL-KL }} & 93.85  \\
\hlineB{4}
\end{tabular}
\caption{A comparison of retrieved contexts and document-level contexts. ${\dagger}$: These approaches are trained on training and development sets.}
\label{tab:document}
\end{table}

\section{Retrieved Contexts Versus Document-level contexts on CoNLL-03}
\label{app:versus}
We conduct a comparison between our retrieved contexts and the document-level contexts on CoNLL-03 datasets. In Table \ref{tab:document}, we report the best model on development set following \citet{yamada-etal-2020-luke}. Comparing with previous state-of-the-art approaches with encoding document-level contexts, our approaches are competitive and even stronger than some of the previous approaches utilizing maximal document-level contexts. Comparing with our model trained on document-level contexts ({\sc\textbf{w/ Doc Context}}), we find that there is still a gap between the document-level contexts and retrieved contexts but our CL approaches can reduce the gap between these two contexts.

\end{document}